**Visual Privacy: Current and Emerging Regulations Around Unconsented Video**

**Analytics in Retail**

Scott N. Pletcher

Purdue Polytechnic Institute

December 4, 2022

**Abstract**

Video analytics is the practice of combining digital video data with machine learning models to infer various characteristics from that video. This capability has been used for years to detect objects, movement and the number of customers in physical retail stores but more complex machine learning models combined with more powerful computing power has unlocked new levels of possibility. Researchers claim it is now possible to infer a whole host of characteristics about an individual using video analytics–such as specific age, ethnicity, health status and emotional state. Moreover, an individual's visual identity can be augmented with information from other data providers to build out a detailed profile–all with the individual unknowingly contributing their physical presence in front of a retail store camera. Some retailers have begun to experiment with this new technology as a way to better know their customers. However, those same early adopters are caught in an evolving legal landscape around privacy and data ownership. This research looks into the current legal landscape and legislation currently in progress around the use of video analytics, specifically in the retail in-store setting. Because the ethical and legal norms around individualized video analytics are still heavily in flux, retailers are urged to adopt a 'wait-and-see' approach or potentially incur costly legal expenses and risk damage to their brand.

*Keywords*: video analytics, computer vision, machine learning, retail stores, privacy laws

**Table of Contents**

**Visual Privacy: Current and Emerging Regulations Around Unconsented Video Analytics in Retail**

Retailers have long used in-store video data for loss prevention and general shopper analytics but recent advances in artificial intelligence-powered video analytics now claim to infer much more specific information from those same video streams, such as a customer's emotional state, health status and even political leaning and sexual preference. With most retailers striving to better personalize the shopping experience, these AI services might provide an efficient and non-invasive way to further build that customer profile. However, for most jurisdictions in the United States, the legal status of collecting unconsented biometric data such as that in video analytics is a gray area with privacy implications. For the retailer, this emerging area presents new opportunities to better know customers but may also put the retailer at risk–both legally and ethically. This research hopes to expand on some clarity in the video analytics space for retailers by reviewing the current legal landscape and emerging trends which might impact the use of video analytics in the retail setting, which seem to be underrepresented in current literature.

**Background**

On April 14, 2017, global retailer Walmart filed a patent application titled "Systems and Methods for Facilitating Shopping in a Physical Retail Facility" (Wilkinson et al., 2020). In this filing, the applicant outlines a vision of retailing that involves instant identification of the individual, matching that individual to existing shopper profile if it exists and augmenting that profile as the person browses around the store. Ultimately the system would provide recommendations to the shopper based on that assembled and derived collection of preferences. While this seems like any other e-Commerce based shopping experience, this experience is different in that it happens in physical stores where shoppers are instantly identified by, among other possible methods, facial recognition.  There is no evidence Walmart has since put any of this into practice but other retailers have, using video data for inferring information about customers (Fight for the Future, 2021).

Video collection devices are prevalent in retail environments, mass transit facilities, sporting venues, and along many streets and highways.  Most of these cameras were installed in the name of safety and theft prevention but modern technology opens up new use cases.  Formerly, analysis and processing of video data was a manual effort, requiring a human to review the footage to extract any value.  Through the enabling technology of computer vision, machine learning and cloud computing, video analysis can now scale infinitely, allowing video data to be processed and inferences made in real-time (Mane & Shah, 2019).  The Supreme Court has maintained "a person does not have a reasonable expectation of privacy with regard to physical characteristics that are constantly exposed to the public, such as one's facial features, voice, and handwriting" (Bowyer, 2004, p. 15).  However, as video processing capabilities increase, enterprising claims have been made that raw video data can now provide a host of other attributes such as existence of physical or mental health issues, emotional state or future job performance despite mixed evidence for their veracity (Harwell, 2022; Le Mau et al., 2021; Luo et al., 2019; Martinez-Martin, 2019).  Using video to infer certain characteristics or predictions of individuals seems to go beyond that which is in public view.  Moreover, these inferences are just best guesses based on mathematical models which identify similarities among pixels in an image rather than based on any factual input from the subject.  Retailers who lean too heavily on such technology may very well be taking on some unexpected and unknown risks.

**Literature Review**

*Emergence of Video Analytics in Retail*

Video surveillance in physical retail store settings has become as ubiquitous as point of sale terminals or weekly sales flyers. Cameras were originally installed for two main reasons: "general protection and surveillance" and "point-specific uses..such as watching to see if certain employees are using drugs…or under the suspicion of theft" (Christine, 1992, p. 14). Additionally, some employers used surveillance video to investigate worker compensation injury claims, potentially debunking fraudulent claims. Beyond the loss prevention and safety concerns, video recordings did not hold much other value for retailers, largely because analysis of the video data required a human to watch the videos.

Technological advances in data collection methods and the ability to transport, store and process massive amounts of structured and unstructured data has created some new possibilities. Modern retailers now collect far more information about employees *and* customers:

> When consumers are inside the store, retailers can: collect their shopping and personal information; capture their images; record their conversations; track, and even deny, their merchandise returns; biometrically surveil their bodies (for example, their faces and fingerprints; and/or track their location inside, and maybe even outside the store. (Elnahla & Neilson, 2021, p. 331)

With all these available data points, retailers are keen on leveraging this information to ultimately drive more profitability through initiatives such as improved customer service, optimized store layouts and more impactful marketing efforts. The act of coelessing all this data into something that can be used to inform decisions or draw conclusions is often referred to as data analytics and a specific subset of that practice that focuses on video data is called *video analysis* (Choudhary & Chaudhury, 2016). Video analytics improves upon this by augmenting video analysis data with other environmental and transactional data to make inferences, synthesize new data and inform decisions.

*Current Use Cases of Video Analytics*

Kröckel & Bodendorf (2012) present a video analytics-based method to track customer movement within the retail store and derive information from those movements. Liciotti et al. (2014) developed a system which uses a series of cameras to track how customers interact with products on the shelves, allowing retailers to optimize shelf layout and where to place additional promotional calls to action. Trinh et al. (2012) enhanced the traditional video-only employee theft prevention use case by adding video analytics combined with point-of-sale transaction log analysis to lower the occurrences of false positives for employee theft allegations.

In recent years, improvements in technology have enabled new use cases. Ijjina et al. (2020) demonstrated the ability to use raw closed-circuit security footage and deep machine learning to identify customers' gender with 82.9% accuracy and age with 70.8% accuracy. Bodziak & Steć (2017) combined in-store sensor data, including customer movement and radio tomography with video analytics with external customer history and social media data to demonstrate retailers can better target marketing messages to the individual–both in-store and online. Other researchers have developed methods to identify and individually track customers within the store by using a combination of video cameras and electro-magnetic emissions of an individual's mobile device, then dynamically push advertisements to their mobile device as so-called 'calls to action' (Liu et al., 2018; Zeng et al., 2019). Group buying behavior can also be derived using video analytics in retail settings (Gandomi & Haider, 2015). While a whole family may enter the store, usually only one person will be involved in the final checkout transaction. Video analytics can provide more granular information about the group, such as quantity and approximate ages of family members and how those other group members respond to product placement or special promotional displays.

Modern video analytics go beyond traditional 'footfall' counters to extract as much information as the technology will permit. In some stores, "mannequins have been fitted with high-resolution video cameras and facial recognition technology that can be used to detect the age, gender, race and facial expressions of customers" (Gregorczuk, 2022, p. 64). Garaus et al. (2021) created a system that used facial recognition to derive a customer's emotional state and present them with custom marketing messages at check-out customized to exploit those inferred emotions. While the idea of using facial recognition to derive demographic information about shoppers is not new, the emergence of cloud-based turn-key facial recognition services has allowed retailers to implement these capabilities without investing in extensive infrastructure or hiring machine learning experts (Guo, 2012; Mane & Shah, 2019).

This ease of implementation can certainly speed the time to market for such capabilities but perception, security and ethical concerns still remain. All the studies reviewed here go into great detail about the technical implementation and the extracted customer data useful to the retailer. However, only one study among the use cases presented here makes any mention of privacy or ethical considerations. Garaus et al. (2021) is this exception, including the ethical considerations and customer perception in their thesis. They found that when customers were told they were being dynamically served up advertisements based on an algorithm and pictures of their face, the customers reported feeling manipulated and discouraged. The authors go on to suggest this as a future area of study, saying "these findings not only point to the relevance of creating legal frameworks on the use of data measurement and processing at the POS, but also raise ethical concerns on targeting practices" (Garaus et al., 2021, p. 758).

Per court filings, retailers Kohl's, The Home Depot, Macy's, Walmart and Best Buy have been named in a suit which alleges the retailers shared in-store video streams with video analytics company Clearview AI (Bilyk & Holland, 2022). Clearview AI provides facial recognition services based on "the largest known database of 30+ billion facial images sourced from public-only web sources, including news media, mugshot websites, public social media and other open sources" (Clearview AI, 2022, para. 2). The filings indicate that these retailers contracted with Clearview AI as a way to individually identify in-store shoppers based on online photographic information Clearview had amassed without the customer's knowledge (*Clearview AI Consumer Privacy Litigation*, 2022).

### *Privacy and Video Analytics*

From the retailer's perspective, these video analytic capabilities serve to fill an important void. Larsen et al. (2017) note that while laboratory-type studies on shopper behavior such as focus groups are somewhat useful, the ability to observe shoppers in-store is by far the more valuable approach to retail analytics. Elnahla & Neilson (2021) found that having an employee, such as a market researcher, directly observing a customer tends to impact that shopper's true behavior–in other words, people will act differently when they know they are being watched. As such, more covert means of observation using video cameras have become popular for conducting in-store research.

Therefore, some retailers consider video analytics as a natural extension of market research–no different or objectionable from a privacy standpoint to the classic 'human-with-a-clipboard' market researcher observing customer behavior and drawing conclusions (Bowers, 1994; Kirkup & Carrigan, 2000). Additionally, market research in general involves an inherent amount of intrusion to one's privacy and can therefore be justified.

However, the legal guidance around video analytics is currently very murky (Elnahla & Neilson, 2021; Fight for the Future, 2021; Norris, 2019; Panahov, 2022). In the absence of such legal clarity, Kirkup & Carrigan (2000) say "the onus falls on the researcher to observe their own personal ethics" including honoring a customer's wish to not participate in the research and thus opt out (p. 474).

When the *researcher* is not a human but instead an artificial intelligence algorithm, this capability to act according to a code of professional or personal ethics does not seem possible. Singh et al. (2017) found that obscuring one's face with a mask, beard, sun glasses or scarf does not ensure one's face cannot still be recorded and identified. As mask mandates of the COVID-19 epidemic became commonplace, facial recognition models have since been optimized to work around face coverings, achieving only a 1% drop in accuracy as compared to unobscured faces (Chen et al., 2022). Even if there was a practical way for a customer to signal non-participation, retailers may not even be required to allow customers to opt-out. While legislation is being considered at the local, state and national level, only Illinois has a law which governs the collection, protection, storage and usage of biometric information such as facial features and fingerprints (Jackson, 2019). While regarded by many as a good starting point, the specific implementation of the law is actively being hashed out through adjudication of suits brought under it.

### *Risks to the Retailer*

Video analytics which use images of the person's face or a digital map of facial features *may* be classified as personally identifiable information or PII. Most retailers are familiar with PII in terms of the Payment Card Industry Data Security Standard (PCI DSS) which governs how retailers need to protect and secure payment card information (Williams & Chuvakin, 2014).

The Health Insurance Portability and Accountability Act (HIPPA) specifically includes such biometric data as personally identifiable information (PII) and thus subject to that act's substantial protections and safeguards (Martinez-Martin, 2019, p. 182).  Additionally, the collector of such PII must inform the customer that this information is being collected and how it will be used.  Online retailers may disclose this in a terms of service document which the customer must agree to as part of a registration process.  However, in a physical retail store, there exists no such gate or explicit authorization.  The collection of such biometric data from in-store customers might not only open up that retailer to the requirements and rigor of HIPPA but may also be problematic from the beginning if the by not securing customer consent.  Several lawsuits have materialized in Illinois and other jurisdictions over technology companies storing and using facial data without first obtaining consent of the subject (*Patel v. Facebook, Inc*, 2019, *Rivera v. Google Inc*, 2017, *Vance v. Microsoft Corporation*, 2021)

Customer sentiment can also be negatively impacted.  Garaus et al. (2021) demonstrated that using facial recognition to derive customer emotions and present advertisements designed to exploit that emotion resulted in customer resentment once the customers were informed. FaceFirst, a vendor of facial recognition systems for retail loss prevention and 'persons of interest' identification specifically iterates "although generally not required by law, we encourage our clients to post signs alerting their patrons when biometric surveillance is being used for public safety purposes" (*Trust*, 2022).

Bias and misidentification in inferences is another risk to the retailer.  Several studies have shown that even the most cutting-edge facial recognition methods still have significant error rates for some segments of the population (Fight for the Future, 2021; Lam, 2019; Le Mau et al., 2021; Mane & Shah, 2019).  Bias among race and ethnicity identification exist in many facial recognition models driven mostly by the underrepresentation of persons of color in raw training data sets (Nieves Delgado, 2022). If these errors are not caught but rather automatically added to a growing customer profile, this not only pollutes the data quality but this, as Custer (2018) notes, "may lead to propagation of existing biases in datasets and resulting patterns, amplifying inequalities and other issues related to profiling even stronger than in regular profiling practices" (p. 4)

Some studies claim to be able to derive ones political orientation, sexual orientation, criminal inclination and genetic disorders from a person's face using video analytics and machine learning techniques (Gurovich et al., 2019; Kosinski, 2021; Wang & Kosinski, 2018; Wu & Zhang, 2019). Other researchers have equated some of these claims to modern 'algorithmic phrenology', referring back to the nineteenth century practice of divining characteristics of a person based on the shape of their head (Aguera y Arcas et al., 2017; Završnik, 2020). While customers would not routinely offer up these personal attributes during an in-store retail transaction, they could nonetheless be used as data points which influence a retailer's behavior and customer's experience. Moreover, with the black box nature of machine learning models, it may be impossible for the retailer to refute claims of discrimination within the process in how those inferences were made (Završnik, 2020).

**Current National and State Laws Governing Video Analytics**

There are currently no national regulations which pertain to the collection and use of video for purposes of inference about the subjects (Rowe, 2020). Although the U.S. government does make use of facial recognition for law enforcement, military actions and border control, there are no regulations in place for its use by government entities. Recently, the government's skill in facial recognition came to light as the technology was used to quickly identify hundreds of individuals within days from video taken during the January 6th raid on the U.S. Capital (Barlow Keener, 2022).

With no guidance or legislation at the federal level, many states have taken up their own effort for privacy legislation.  Illinois was considered an early adopter in legislation on safeguarding biometric information when the Illinois Biometric Privacy Act ("BIPA") was signed into law in 2008 (McMahon, 2021).  In the time since the bill went into place, several cases have been brought under the law with mixed results.  In 2020, California passed the California Privacy Rights Act which classifies biometric data as *sensitive personal information*, requiring companies that collect such information to make their collection known to individuals, maintain certain data retention guidelines and prohibit the sale or transfer of that information (Biometrics Information, 2022).  This act further augmented California's existing Consumer Privacy Act by placing more specific wording around biometric data in particular.

By comparison, Texas and Washington have very similar statutes to Illinois BIPA in terms of use and collection of biometric data but the Texas and Washington laws differ in a key aspect–neither statute provides for the ability of a private citizen to bring cause of action against companies suspected of violating the terms (Browning, 2018).  Rather, the respective state's attorney general must take up the cause, resulting in very few cases brought forward given other more pressing and higher visibility matters.  Washington's law does offer perhaps the broadest definition of biometric data by defining it as "any data generated by automatic measurements of an individual's biological characteristics" (Browning, 2018, p. 676).  In 2020, Vermont passed a moratorium on facial recognition and uniquely prohibits not only identification but also the ability to "determine the person's sentiment, state of mind, or other propensities" through video analytics (Moratorium on Facial Recognition Technology, 2020, sec. 14(b)(1)(B)).  However, this moratorium only applies to law enforcement and does not cover any facial recognition in commercial or private scenarios (Moratorium on Facial Recognition Technology, 2020). Virginia, Maryland, New Hampshire, New York, Oregon, Maine and Utah all have similar bills in various stages of legislation (Barlow Keener, 2022).

This patchwork of legislation around video analytics creates a very challenging landscape for retailers operating on a national level. In an effort to simplify and better define the legal picture, many companies have become some of the most vocal advocates for national regulation (Rowe, 2020). Other companies who find themselves as defendants in cases brought under these existing state privacy laws have actively sought to move the case into the Federal courts (*Clearview AI Consumer Privacy Litigation*, 2022, *Patel v. Facebook, Inc*, 2019, *Rivera v. Google Inc*, 2017).

### *Expectations of Privacy in Retail Spaces*

The Supreme Court has maintained "a person does not have a reasonable expectation of privacy with regard to physical characteristics that are constantly exposed to the public, such as one's facial features, voice, and handwriting" (Bowyer, 2004, p. 15). So long as cameras are not placed in dressing rooms, bathrooms or other areas where people *do* have expectation of privacy, retailers are well within their and their customers' rights to capture video footage in and around their physical stores. However, much of this interpretation was formed long before current computer vision and video analytics capabilities. Because machine learning video analytic models are optimized to pick up very subtle characteristics, they can detect potential attributes which humans viewing the same video cannot. As discussed earlier, current technology allows inference of many attributes from video feeds which customers may not care to overtly disclose–such as health status or emotional sentiment. In Kyllo v. United States (2001), the Supreme Court ruled that it was a violation of the Fourth Amendment when law enforcement used a thermal imaging camera to effectively search a private home from a public space without a warrant.  One might see some parallels with a thermal imaging camera and modern AI-enabled video cameras, especially if those cameras and subsequent image processing are still able to infer identifying information despite a customer's efforts to conceal their identity as Singh et al. (2017) and Chen (2022) have demonstrated.

*Machine Learning Algorithm Usage and Inference Ownership*

According to consumer protection group Fight for the Future (2021), retailers have currently taken up different positions on use of video inference within stores.  Walmart, Target, Lowes, Kroger and Home Depot are among retailers who have gone on record as not using facial recognition.  Conversely, Albersons and Macy's affirm that they use such technology via company statements and privacy policies (Fight for the Future, 2021; *Privacy Policy*, 2022).  A host of other retailers have not made explicit statements either way on their use of video analytics for individual inferences.

In cases where facial recognition systems were alleged to be used in retail settings, the defendants have gone to great lengths to protect the specific details of how their systems work. In the pending class action lawsuit against Clearview AI, a facial recognition services vendor, the company has repeatedly rebuffed attempts in the discovery process to allow third-party expert witnesses to review its video analytics algorithms, claiming they were trade secrets (*Clearview AI Consumer Privacy Litigation*, 2022).  In 2020, Clearview AI suffered a data breach in which its client list was leaked (Morrison, 2020).  Although Clearview AI claims its service is intended to serve the law enforcement community, several retail companies appeared on the client list, including Walmart, Best Buy, Albertsons, Rite Aid, Kohls and Macy's.  When asked for comment, both Walmart and Best Buy claim that they were not Clearview clients at the time of the breach and that they only conducted a brief trial of the service before declining further use.

In *Pruitt v. Par-A-Dice Casino* (2020), a casino claimed their video surveillance system was not even capable of facial recognition and insisted that the plaintiff had no basis for making such a claim, asking the court to dismiss the case all together. The court refused to dismiss the case on the grounds that it was indeed plausible that a casino would have such biometric collection capabilities and that the discovery process would bear that out as true or false. In most cases filed under Illinois' BIPA law, the issue is not whether video analytics are taking place but rather was biometric data collected without consent.

Even if there is consent secured from patrons, other questions arise as to the ownership of the inferences made using that biometric data. Many companies have maintained that any data synthesized by them about customers is just more knowledge they have developed to augment existing customer profile data (Wrabetz, 2022). Additionally, some companies also consider inferred data as intellectual property or trade secrets and therefore should not have to be disclosed as it could be detrimental to the company's business. This stance is at odds with most of the state privacy laws in place or in process, as many include provisions for customers to know exactly what data a company holds about them, the ability to verify accuracy or require the company to delete their data. Additionally, inferences are creations of computer algorithms rather than humans and "under current U.S. copyright and U.S. patent law, a work can only be an original work of authorship or invention respectively if it was created by a human being" (Wrabetz, 2022, para. 5).

The trade secrets argument seems less plausible if there is no real *secret recipe* which the company uses to uniquely extract these inferences from their video feeds. Many software-as-a-service providers offer turn-key video analytic and facial recognition capabilities, allowing companies to outsource the technical analysis of the video feed and get back attributes or characteristics. In exchange for simplicity and cost-effectiveness, companies accept these services as *black box* implementations without regard for the inner workings of the models. It might be hard to persuade a court that some special proprietary process was used to generate the new inferred data if the inferences were achieved using service available to any competitor with a credit card. Additionally, if the biometric data is considered sensitive personal information, then any derivative information should also be classified the same and thus subject to the customer rights for consent, accuracy, use and removal (Wrabetz, 2022).

**Emerging Regulations on Video Analytics in the United States**

Beyond the current laws, several other legislative efforts are underway which could ultimately impact the use of video analytics in the retail setting. During the 2021-2022 Congressional session, 35 bills have been introduced which propose some form of data privacy. According to legal and privacy scholars, the most promising bill seems to be the American Data Privacy and Protection Act (ADPPA) introduced by Representative Frank Pallone Jr. (ADPPA, 2022; Blanke, 2022; Hartzog & Richards, 2022). Hartzog & Richards (2022) call the bill "the most significant bipartisan privacy legislation introduced in more than a decade", going on describe it as "a sincere attempt to move beyond the ineffective 'notice and choice' approach to privacy that has been the hallmark of U.S. legislators since the days of dial-up modems" (2022, para. 3).

*American Data Privacy and Protection Act*

The ADPPA does cover biometric data, specifically including facial mappings and "gait or personally identifying physical movements" which would typically be used for individualized video analytics . The wording also excludes video recording from being considered biometric information as well as any information which could be gleaned from a video recording so long as that data is not used to specifically identify an individual. For video analytics toward an individual, biometric information could still be collected but such data would be considered sensitive and subject to a host of requirements around its collection, transportation, storage and destruction. Additionally, the collecting entity must ensure the customer is aware of the collection, most likely through an explicit acknowledgement, and maintain transparency around the use of the information (ADPPA, 2022, sec. 201). If the individual does not consent to providing the sensitive data, the entity collecting the data may not refuse or alter the product or service based on that person's decision. Loyalty programs would still be allowed as would providing discounts or extra benefits in exchange for continued patronage so long as the customer consented to providing any sensitive data which might be required for enrolling in the program. Current loyalty programs in which the customer identifies themselves at checkout by providing a customer number or phone number still fit within this bill. However, using video feeds and machine learning inferences as a way to build individual profiles of normally anonymous shoppers and then provide them with customized promotions or a different customer experience does not seem to fit within the text of the bill.

      As currently written, the ADPPA includes a specific clause which allows this federal bill to preempt state laws covering the same provisions. For national retailers, this might seem to be a welcome addition as it could create some commonality from state to state. Unfortunately for national retailers, that same section also includes specific exemptions of preempting Illinois' BIPA and "laws that solely address facial recognition or facial recognition technologies" (ADPPA, 2022, sec. 404(b)(2)(K)). With this clause, states would be free to craft their own laws

around such technology which, for video analytics seeking to infer characteristics about an individual, does nothing to progress a consistent federal law.  Additionally, states push back, wanting to preserve more autonomy.  Blanke (2022) cites the example of the federal CAN-SPAM Act of 2003, which by the time it eventually progressed to Congressional vote, nearly every state had already passed their own much stronger anti-spam laws.  Because the federal anti-spam bill was far weaker than the respective state versions, 44 state attorney generals refused to support the bill.

### *Blueprint for an AI Bill of Rights*

In October, 2022, the Office of Science and Technology Policy (OSTP) published the "Blueprint for an AI Bill of Rights", a document which describes a "set of five principles and associated practices to help guide the design, use, and deployment of automated systems to protect the rights of the American public in the age of artificial intelligence" (OSTP, 2022, p. 4).  While not legally binding, the document does signal Executive branch acknowledgement that AI is transformative and brings new responsibilities.  This document shares many of the same key recommendations and practices as various other privacy bills and laws such as the expectation that private data remain private and that the individual should maintain ownership of their data.  When AI processes are used which impact the individual, the individual should know that AI is being used and be able to understand how and why that AI process had influence on a given outcome.  Perhaps the most weighty topic included in the AI Bill of Rights is the expectation of the individual to not encounter algorithmic discrimination.  As discussed earlier, bias can exist in machine learning algorithms and is exacerbated by the lack of explainability.

*Common Privacy Themes in Emerging Legislation*

Although the current generation of privacy laws in place and in process have some differences, common themes have emerged at their core and any retailer crafting a forward-looking strategy should take notice. Disclosure and consent are at the heart of most bills, ensuring that customers are aware that personal data, including individualized video analytics and biometric data, is being collected and used. If customers do not wish to provide that consent, there must be a customer experience path which does not require such data to be collected and not materially different or curtailed compared to the consented customer path. Whether or not customer consent is granted, some degree of fear, disapproval or suspicion will likely negatively impact brand reputation until such practices become more common and transparent.

**Implications for Retailer Strategy on Individualized Video Analytics**

Retailers should take away three points on the topic of in-store video analytics. First, being a first-mover in individualized video analytics may not be a wise strategy. More legal precedent is needed to test and shape the emerging privacy laws and retailers who launch aggressive video analytics efforts now may find themselves quickly on the wrong side of new laws and public opinion. Second, if retailers choose to profile individuals with video analytics, they should disclose this activity in the name of transparency and openness for their customers and certainly not try to hide behind the current absence of a legal requirement. The dilemma for the retailer employing individualized video analytics is in how to disclose this to customers without invoking *Orwellian* fears.

Finally, retailers must clearly demonstrate the benefit *to the customer* of collecting and using this biometric information. Just as with loyalty programs, there must be some perceived value in the minds of customers and not just a one-sided method to subject customers to more advertising.  One area for further study would be what new features or benefits from retail video analytics might constitute a proper *quid pro quo* in the minds of customers.  Individualized video analytic technology does hold the potential to enable more efficient, frictionless retail experiences, but current technology and legislation are limiting factors.

**Limitations**

This research has intentionally focused solely on in-store retail scenarios and specifically excluded other common and current video analytic use cases.  Law enforcement has made use of video analytics and facial recognition throughout  the United States for years to search crowds for people of interest and to read license plates (Bowyer, 2004).  In general, public opinion around video analytics for safety and security tends to be more sympathetic as compared to video analytics for commercial reasons.  While there is certainly a share of controversy around the use of video analytics for law enforcement, it was important to not commingle these different use cases.

**Conclusion**

Machine learning and digital video imagery has and will continue to enable incredible capabilities.  However, just like any other technology maturity S-curve, this technology is currently in an early steep phase of rapid innovation which is helping define what *can* be done.  To evolve into the next phases of adoption, what *should* be done must also be considered and adopted as norms.  Modern retailers would be wise to adopt a wait-and-see approach as the first-movers in this space are already bearing the brunt of this legal evolution in the courts.  As this technology is already commoditized, being a fast-follower once more clarity exists seems like the path of least risk. Alternatively, rather than eagerly jumping on the video analytics bandwagon, perhaps the best way to establish competitive differentiation is to become the most privacy-focused retailer in an industry obsessed with spinning that next morsel of data into gold.

# References

Aguera y Arcas, B., Mitchell, M., & Todorov, A. (2017, May 7). *Physiognomy's New Clothes*. Medium. https://medium.com/@blaisea/physiognomys-new-clothes-f2d4b59fdd6a

American Data Privacy and Protection Act, 8152, House of Representatives, 117th Congress (2022). https://www.congress.gov/bill/117th-congress/house-bill/8152?q=%7B%22search%22%3A%5B%228152%22%2C%228152%22%5D%7D&s=10&r=3

Barlow Keener, E. (2022, April 25). *Facial Recognition: A New Trend in State Regulation*. American Bar Association. https://www.americanbar.org/groups/business_law/publications/blt/2022/05/facial-recognition/

Bilyk, J., & Holland, S. (2022, August 15). *Biometrics face scan class actions filed vs Walmart, Kohls, Best Buy, Home Depot, over involvement with Clearview*. Cook County Record. https://cookcountyrecord.com/stories/630173081-biometrics-face-scan-class-actions-filed-vs-walmart-kohls-best-buy-home-depot-over-involvement-with-clearview

Biometrics Information, SB-1189, California Legislature, 2021-2022 Regular Session (2022). https://leginfo.legislature.ca.gov/faces/billTextClient.xhtml?bill_id=202120220SB1189

Blanke, J. (2022). The CCPA, "Inferences Drawn," and Federal Preemption. *Richmond Journal of Law and Technology*, *29*(1). https://doi.org/10.2139/ssrn.4219967

Bodziak, B., & Steć, B. (2017). The acquisition of consumer behaviour data using integrated indoor positioning systems. *Organization & Management Scientific Quartely*, *nr 3*. https://doi.org/10.29119/1899-6116.2017.39.2

Bowers, D. K. (1994). Privacy concerns and the research industry. *Marketing Research; Chicago*, *6*(2), 48–50. https://www.proquest.com/scholarly-journals/privacy-concerns-research-industry/docview/202712779/se-2

Bowyer, K. W. (2004). Face recognition technology: security versus privacy. *IEEE Technology and*

*Society Magazine*, *23*(1), 9–19. https://doi.org/10.1109/MTAS.2004.1273467

Browning, J. (2018). The Battle over Biometrics. *Texas Bar Journal*, *81*(674). https://advance-lexis-com.ezproxy.lib.purdue.edu/api/document?collection=analytical-materials&id=urn:contentItem:5TJX-FWJ0-01XY-P34Y-00000-00&context=1516831

Chen, H.-Q., Xie, K., Li, M.-R., Wen, C., & He, J.-B. (2022). Face Recognition With Masks Based on Spatial Fine-Grained Frequency Domain Broadening. *IEEE Access*, *10*, 75536–75548. https://doi.org/10.1109/ACCESS.2022.3191113

Choudhary, A., & Chaudhury, S. (2016). Video analytics revisited. *IET Computer Vision*, *10*(4), 237–249. https://doi.org/10.1049/iet-cvi.2015.0321

Christine, B. (1992). New tricks for video surveillance. *Risk Management*, *39*(8), 14–16. https://www.ncbi.nlm.nih.gov/pubmed/10120622

Clearview AI. (2022). *Overview*. Clearview AI. https://www.clearview.ai/overview

Clearview AI Consumer Privacy Litigation, 585 F.Supp.3d ___ (N.D. Ill. 2022). https://scholar.google.com/scholar_case?case=2094407183176999695&q=Walmart+BIPA&hl=en&as_sdt=800006

Custers, B. (2018). Profiling as inferred data. Amplifier effects and positive feedback loops. In E. Bayamlioğlu, I. Baraliuc, L. Janssens, & M. Hildebrandt} (Eds.), *Being Profiled: 10 Years of Profiling the European Citizen*. Elsevier BV. https://doi.org/10.2139/ssrn.3466857

Elnahla, N., & Neilson, L. C. (2021). Retaillance: a conceptual framework and review of surveillance in retail. *The International Review of Retail, Distribution and Consumer Research*, *31*(3), 330–357. https://doi.org/10.1080/09593969.2021.1873817

Fight for the Future. (2021). *Ban Facial Recognition in Stores*. Ban Facial Recognition in Stores. https://www.banfacialrecognition.com/stores/

Gandomi, A., & Haider, M. (2015). Beyond the hype: Big data concepts, methods, and analytics. *International Journal of Information Management*, *35*(2), 137–144. https://doi.org/10.1016/j.ijinfomgt.2014.10.007

Garaus, M., Wagner, U., & Rainer, R. C. (2021). Emotional targeting using digital signage systems and facial recognition at the point-of-sale. *Journal of Business Research*, *131*, 747–762. https://doi.org/10.1016/j.jbusres.2020.10.065

Gregorczuk, H. (2022). Retail Analytics: Smart-Stores Saving Bricks and Mortar Retail or a Privacy Problem? *Law, Technology and Humans*, *4*(1), 63–78. https://lthj.qut.edu.au/article/view/2088

Guo, G. (2012). Human Age Estimation and Sex Classification. In C. Shan, F. Porikli, T. Xiang, & S. Gong (Eds.), *Video Analytics for Business Intelligence* (pp. 101–131). Springer Berlin Heidelberg. https://doi.org/10.1007/978-3-642-28598-1_4

Gurovich, Y., Hanani, Y., Bar, O., Nadav, G., Fleischer, N., Gelbman, D., Basel-Salmon, L., Krawitz, P. M., Kamphausen, S. B., Zenker, M., Bird, L. M., & Gripp, K. W. (2019). Identifying facial phenotypes of genetic disorders using deep learning. *Nature Medicine*, *25*(1), 60–64. https://doi.org/10.1038/s41591-018-0279-0

Hartzog, W., & Richards, N. (2022). Professors Hartzog and Richards Advocate for Data Loyalty in Privacy Legislation. *Technology Academics Policy (TAP)*, *July 25, 2022*. https://scholarship.law.bu.edu/shorter_works/169/

Harwell, D. (2022). A face-scanning algorithm increasingly decides whether you deserve the job. In *Ethics of Data and Analytics* (pp. 206–211). Auerbach Publications. https://doi.org/10.1201/9781003278290-31

Ijjina, E. P., Kanahasabai, G., & Joshi, A. S. (2020). Deep Learning based approach to detect Customer Age, Gender and Expression in Surveillance Video. *2020 11th International Conference on Computing, Communication and Networking Technologies (ICCCNT)*, 1–6. https://doi.org/10.1109/ICCCNT49239.2020.9225459

Jackson, M. (2019). Opting out: Biometric information privacy and standing. *Duke Law and Technology Review*, *18*, 293. https://heinonline.org/hol-cgi-bin/get_pdf.cgi?handle=hein.journals/dltr18§ion=21

Kirkup, M., & Carrigan, M. (2000). Video surveillance research in retailing: ethical issues. *International*

VIDEO ANALYTICS REGULATIONS AND RETAIL	26*Journal of Retail & Distribution Management; Bradford*, *28*(11), 470–480.

https://doi.org/10.1108/09590550010356831

Kosinski, M. (2021). Facial recognition technology can expose political orientation from naturalistic facial images. *Scientific Reports*, *11*(1), 1–7. https://doi.org/10.1038/s41598-020-79310-1

Kröckel, J., & Bodendorf, F. (2012). Customer Tracking and Tracing Data as a Basis for Service Innovations at the Point of Sale. *2012 Annual SRII Global Conference*, 691–696. https://doi.org/10.1109/SRII.2012.115

Kyllo v. United States, 533 U.S. 27 (U.S. 2001). https://scholar.google.com/scholar_case?case=15840045591115721227

Lam, K. (2019, January 30). *Incident 288*. Artificial Intelligence Incident Database; Responsible AI Collaborative. https://incidentdatabase.ai/cite/288

Larsen, N. M., Sigurdsson, V., & Breivik, J. (2017). The Use of Observational Technology to Study In-Store Behavior: Consumer Choice, Video Surveillance, and Retail Analytics. *The Behavior Analyst / MABA*, *40*(2), 343–371. https://doi.org/10.1007/s40614-017-0121-x

Le Mau, T., Hoemann, K., Lyons, S. H., Fugate, J. M. B., Brown, E. N., Gendron, M., & Barrett, L. F. (2021). Professional actors demonstrate variability, not stereotypical expressions, when portraying emotional states in photographs. *Nature Communications*, *12*(1), 5037. https://doi.org/10.1038/s41467-021-25352-6

Liciotti, D., Contigiani, M., Frontoni, E., Mancini, A., Zingaretti, P., & Placidi, V. (2014). Shopper Analytics: A Customer Activity Recognition System Using a Distributed RGB-D Camera Network. *Video Analytics for Audience Measurement*, 146–157. https://doi.org/10.1007/978-3-319-12811-5_11

Liu, X., Jiang, Y., Jain, P., & Kim, K.-H. (2018). TAR: Enabling Fine-Grained Targeted Advertising in Retail Stores. *Proceedings of the 16th Annual International Conference on Mobile Systems, Applications, and Services*, 323–336. https://doi.org/10.1145/3210240.3210342

Luo, H., Yang, D., Barszczyk, A., Vempala, N., Wei, J., Wu, S. J., Zheng, P. P., Fu, G., Lee, K., & Feng, Z.-P. (2019). Smartphone-Based Blood Pressure Measurement Using Transdermal Optical Imaging

Technology. *Circulation. Cardiovascular Imaging*, *12*(8), e008857.

https://doi.org/10.1161/CIRCIMAGING.119.008857

Mane, S., & Shah, G. (2019). Facial Recognition, Expression Recognition, and Gender Identification. *Data Management, Analytics and Innovation*, 275–290.

https://doi.org/10.1007/978-981-13-1402-5_21

Martinez-Martin, N. (2019). What Are Important Ethical Implications of Using Facial Recognition Technology in Health Care? *AMA Journal of Ethics*, *21*(2), E180–E187.

https://doi.org/10.1001/amajethics.2019.180

McMahon, M. (2021). *Illinois Biometric Information Privacy Act Litigation in Federal Courts: Evaluating the Standing Doctrine in Privacy Contexts*. https://papers.ssrn.com/abstract=3929645

Moratorium on Facial Recognition Technology, no. S.124, Vermont Senate (2020).

https://legislature.vermont.gov/Documents/2020/Docs/BILLS/S-0124/S-0124%20As%20Passed%20by%20Both%20House%20and%20Senate%20Unofficial.pdf

Morrison, S. (2020, February 26). *The world's scariest facial recognition company is now linked to everybody from ICE to Macy's*. Vox.com.

https://www.vox.com/recode/2020/2/26/21154606/clearview-ai-data-breach

Nieves Delgado, A. (2022). Race and statistics in facial recognition: Producing types, physical attributes, and genealogies. *Social Studies of Science*, 3063127221127666.

https://doi.org/10.1177/03063127221127666

Norris, S. (2019). And the Eye in the Sky Is Watching Us All-The Privacy Concerns of Emerging Technological Advances in Casino Player Tracking. *UNLV Gaming LJ*, *9*, 269.

https://heinonline.org/hol-cgi-bin/get_pdf.cgi?handle=hein.journals/unlvgalj9§ion=21

OSTP. (2022). *Blueprint for an AI Bill of Rights: Making Automated Systems Work for the American People*. The White House.

https://www.whitehouse.gov/wp-content/uploads/2022/10/Blueprint-for-an-AI-Bill-of-Rights.pdf

Panahov, H. (2022). On Facial Recognition Technology. *Intersect: The Stanford Journal of Science,*

*Technology, and Society*, *15*(2). https://ojs.stanford.edu/ojs/index.php/intersect/article/view/2168

Patel v. Facebook, Inc, 932 F.3d 1264 (9th Cir. 2019).

    https://scholar.google.com/scholar_case?case=9033020751616130750&q=patel+v+facebook&hl=en&as_sdt=800006

*Privacy Policy*. (2022). Albertsons Companies, Inc.

    https://albertsonscompanies.com/policies-and-disclosures/privacy-policy/default.aspx

Pruitt v. Par-A-Dice Hotel Casino, No. 1:20-CV-01084-JES-JEH (Dist. Court, CD Illinois August 31, 2020). https://scholar.google.com/scholar?scidkt=12380548637727511437&as_sdt=2&hl=en

Rivera v. Google Inc, 238 F.Supp.3d 1088 (N.D. Ill. 2017).

    https://scholar.google.com/scholar_case?case=14457227830664707660&q=Rivera+vs+Google&hl=en&as_sdt=800006

Rowe, E. A. (2020). *Regulating Facial Recognition Technology in the Private Sector*.

    https://papers.ssrn.com/abstract=3769613

Singh, A., Patil, D., Meghana Reddy, G., & Omkar, S. N. (2017). Disguised Face Identification (DFI) with Facial KeyPoints using Spatial Fusion Convolutional Network. In *arXiv [cs.CV]*. arXiv. http://arxiv.org/abs/1708.09317

Trinh, H., Pankanti, S., & Fan, Q. (2012). Multimodal ranking for non-compliance detection in retail surveillance. *2012 IEEE Workshop on the Applications of Computer Vision (WACV)*, 241–246. https://doi.org/10.1109/WACV.2012.6163010

*Trust*. (2022). FaceFirst. https://www.facefirst.com/trust

Vance v. Microsoft Corporation, 525 F.Supp.3d 1287 (W.D. Wash. 2021).

    https://scholar.google.com/scholar_case?case=8534915895914540683&q=vance+v+microsoft&hl=en&as_sdt=800006

Wang, Y., & Kosinski, M. (2018). Deep neural networks are more accurate than humans at detecting sexual orientation from facial images. *Journal of Personality and Social Psychology*, *114*(2), 246–257. https://doi.org/10.1037/pspa0000098

Wilkinson, B. W., Jones, M. A., Vasgaard, A. J., Taylor, R. J., Webb, T. W., Mattingly, T. D., & Todd, J. R. (2020). Systems and methods for facilitating shopping in a physical retail facility (USPTO Patent No. 10592959). In *US Patent* (No. 10592959). https://patentimages.storage.googleapis.com/38/c4/69/0c0b602d879902/US10592959.pdf

Williams, B. R., & Chuvakin, A. (2014). *PCI Compliance: Understand and Implement Effective PCI Data Security Standard Compliance*. Syngress. https://play.google.com/store/books/details?id=aWpzAwAAQBAJ

Wrabetz, J. (2022, August 22). *What Is Inferred Data and Why Is It Important?* https://www.americanbar.org/groups/business_law/publications/blt/2022/09/inferred-data/

Wu, X., & Zhang, X. (2019). *Automated Inference on Criminality using Face Images. arXiv (2016)*.

Završnik, A. (2020). Criminal justice, artificial intelligence systems, and human rights. *ERA Forum*, *20*(4), 567–583. https://doi.org/10.1007/s12027-020-00602-0

Zeng, J., Wang, P., Zhao, Q., Pang, J., Tao, J., & Guan, X. (2019). Effectively Linking Persons on Cameras and Mobile Devices on Networks. *IEEE Internet Computing*, *23*(4), 18–26. https://doi.org/10.1109/MIC.2019.2923189